\documentclass[lettersize,journal]{IEEEtran}
\usepackage{amsmath,amsfonts}
\usepackage{algorithmic}
\usepackage{algorithm}
\usepackage{array}
\usepackage[caption=false,font=normalsize,labelfont=sf,textfont=sf]{subfig}
\usepackage{textcomp}
\usepackage{stfloats}
\usepackage{url}
\usepackage{verbatim}
\usepackage{graphicx}
\usepackage{cite}
\usepackage{booktabs}
\begin{document}

\title{Few-shot Image Generation via Style Adaptation and Content Preservation}

\author{Xiaosheng He, Fan Yang, Fayao Liu, Guosheng Lin
        % <-this % stops a space
\thanks{Corresponding author: Guosheng Lin.}
\thanks{ Xiaosheng He, Fan Yang and Guosheng Lin are with School of Computer Science and
Engineering, Nanyang Technological University (NTU), Singapore 639798(email: xiaosheng.he@ntu.edu.sg, fan007@e.ntu.edu.sg,
gslin@ntu.edu.sg)
}% <-this % stops a space
\thanks{Fayao Liu is with Agency for Science, Technology and Research (A*STAR), Singapore 138632 (email: fayaoliu@gmail.com)}}
\markboth{Journal of \LaTeX\ Class Files,~Vol.~14, No.~8, August~2021}%
{Shell \MakeLowercase{\textit{et al.}}: A Sample Article Using IEEEtran.cls for IEEE Journals}


\maketitle

\begin{abstract}
Training a generative model with limited data (e.g., 10) is a very challenging task. Many works propose to fine-tune a pre-trained GAN model. However, this can easily result in overfitting. In other words, they manage to adapt the style but fail to preserve the content, where \textit{style} denotes the specific properties that defines a domain while \textit{content} denotes the domain-irrelevant information that represents diversity. Recent works try to maintain a pre-defined correspondence to preserve the content, however, the diversity is still not enough and it may affect style adaptation. In this work, we propose a paired image reconstruction approach for content preservation. We propose to introduce an image translation module to GAN transferring, where the module teaches the generator to separate style and content, and the generator provides training data to the translation module in return. Qualitative and quantitative experiments show that our method consistently surpasses the state-of-the-art methods in few shot setting.
\end{abstract}


\section{Introduction}
\begin{figure}[t]
    \includegraphics[width= \linewidth]{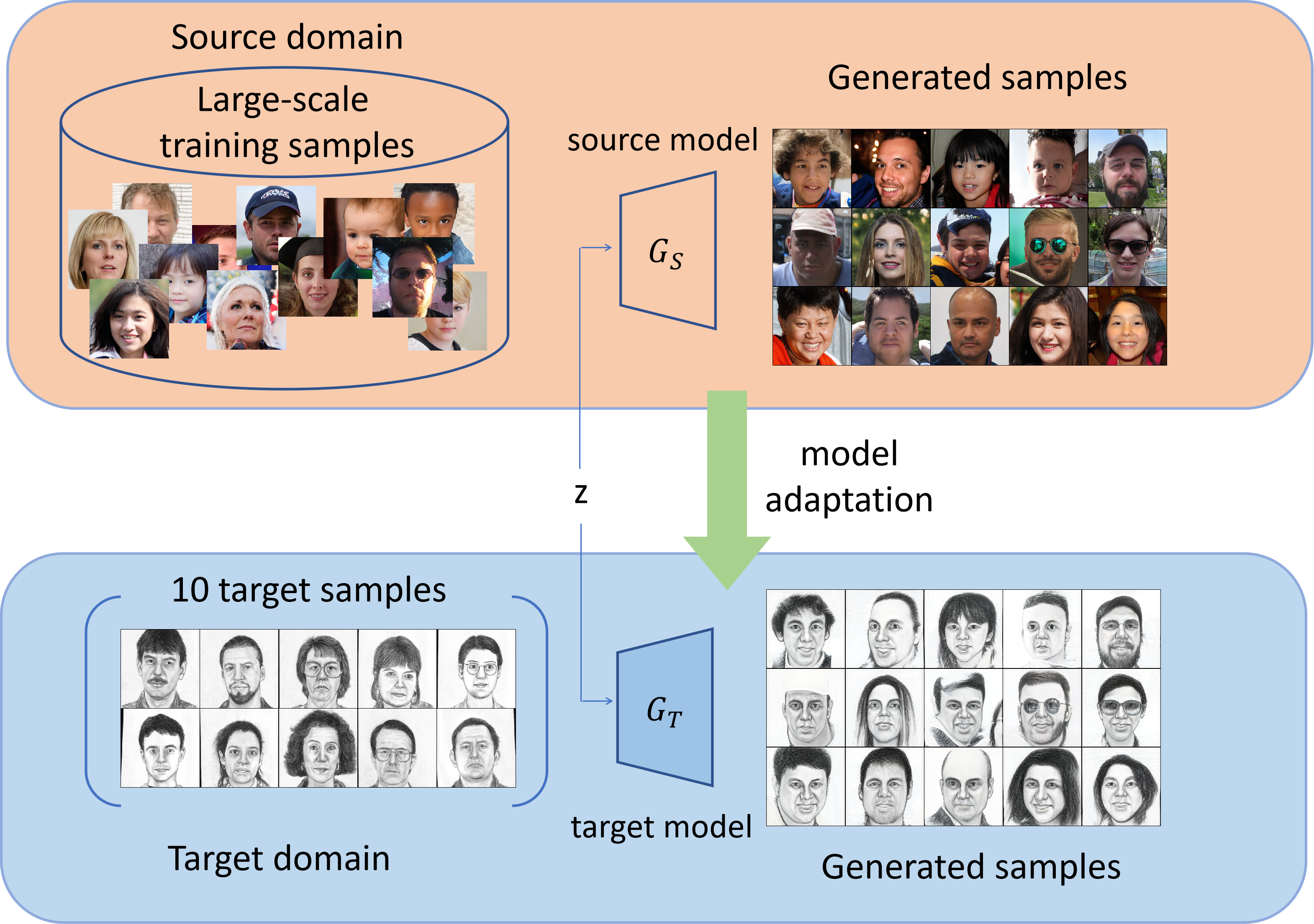} \\
  \caption
    { Given a source model $G_S$ trained on a large-scale dataset, we want to adapt it to a target domain with very few examples. The target model is expected to generate diverse images with the target style.}
    \label{lp}
  \label{fig:1}
\end{figure}

\IEEEPARstart{G}{enerative} adversarial networks (GANs) learn to map a simple pre-defined distribution to a complex real image distribution. Despite its great success in many areas of computer vision including image manipulation \cite{diao2022zergan}, image-to-image-translation \cite{liu2019few,zhu2017unpaired,tang2021attentiongan,kong2023unpaired,luo2022slogan} and anomaly detection \cite{huang2022self}, GAN requires a large amount of training data and time to achieve high-quality images. Therefore, few-shot generative model adaptation has been proposed, which aims to transfer a pre-trained source generative model to a target domain with extremely limited examples (e.g. 10 images), as is shown in Fig \ref{fig:1}. The practical importance of this task is two-fold: 1. In some domains such as painting, it is very difficult to obtain enough data to meet the training requirements of GANs. 2. a well-trained GAN holds a wealth of knowledge about images, which can be leveraged to train GANs in similar domains and significantly reduce the training time (from weeks to a few hours).

To achieve this, fine-tuning based methods have been proposed, where people only fine-tune a part of the model parameters or train a few additional parameters\cite{noguchi2019image,robb2020few,zhao2020leveraging,wang2020minegan}. Most of these methods, however, still requires hundreds of training images. When the target samples is limited to 10, they are prone to overfitting and fail to inherit the diversity from source domain.

 To address these issues, recent methods tried to constrain the transfer process based on some assumed correspondence between two images. Ojha et al. \cite{ojha2021few} proposed to preserve the differences of relative similarities between instances via cross-domain correspondence (CDC) loss and a patch discriminator. Xiao et al. \cite{xiao2022few} proposed a relaxed spatial structural alignment (RSSA) method and tried to project the original latent space to a narrow subspace close to the target domain to accelerate training. These methods can generate diverse and realistic images with limited data. However, these pre-defined correspondence losses are either relatively weak in diversity preservation (CDC) or overemphasize diversity (RSSA) and sacrifice style adaptation, limiting the gap between these two domains (Fig  \ref{fig:trade}, \ref{fig:sketch}, \ref{fig:spaniel_vangogh}).

\begin{figure}[t]
    {\includegraphics[width=\linewidth]{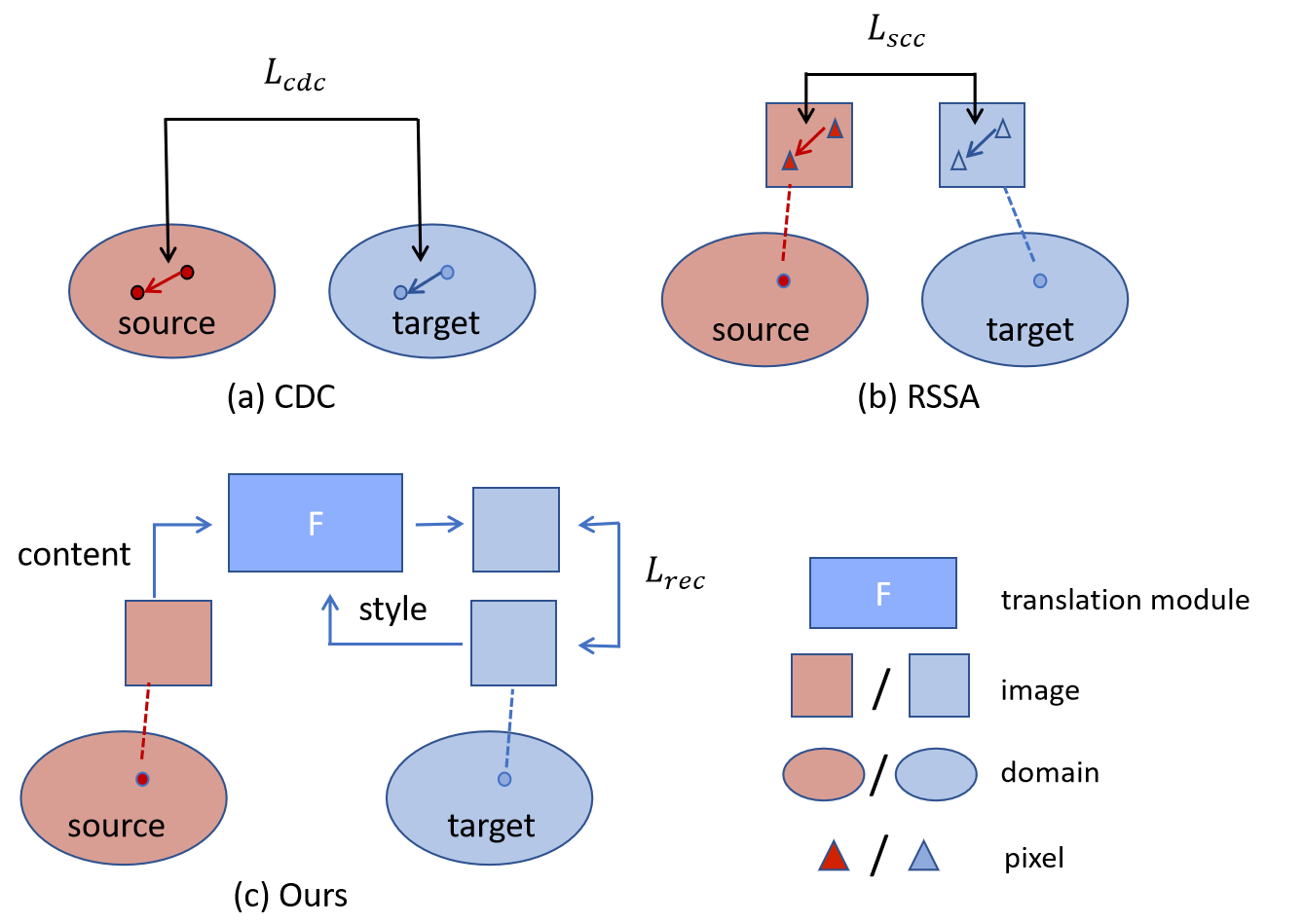}}
  \caption
    { What to preserve in different methods. (a) CDC: preserve the distances between instances. (b) RSSA: preserve the distances between pixels. (c) Our method: preserve content information learned from the translation module $F$. The comparison results are shown in Fig \ref{fig:sketch},  \ref{fig:spaniel_vangogh}}
  \label{fig:method_compare}
\end{figure}

 In this work, we propose PIR, a paired image reconstruction method to address the few-shot generative model adaptation. We first make an assumption that given the same latent code, the source model and the target model should generate a pair of images with the same \textit{content} and different \textit{style}, where \textit{style} denotes the specific properties that defines a domain while \textit{content} denotes the domain-irrelevant information that represents diversity. As is shown in Fig \ref{fig:method_compare}, the correspondences preserved by CDC and RSSA can be seen as estimations of the real content. Motivated by previous work of style transfer \cite{huang2017arbitrary} and image-to-image translation \cite{liu2019few}, we introduce an image translation module that learns to separate style and content of an image during training (Note that it is still impossible to train a suitable translator
using only data from the source domain and 10 examples from the target domain, as depicted in Fig \ref{fig:sketch}). In the training process, the source and the target model generate a pair of images with the same latent code, we let the translation module to reconstruct them with their own style and each other's content. This reconstruction will thus encourage the adapted generator to inherit the diverse content from the source domain. Instead of applying pre-defined correspondence losses, the model will dynamically balance style adaptation and content preservation to generate realistic and diverse outputs.  
\\ \hspace*{\fill} \\
 \noindent
 \textbf{Contributions.} Our main contribution is a novel paired image reconstruction method to transfer diversity from source domain to target domain. Qualitative and quantitative results shows that our method produces best results in a variety of settings.

\begin{figure}[t]
    \includegraphics[width= \linewidth]{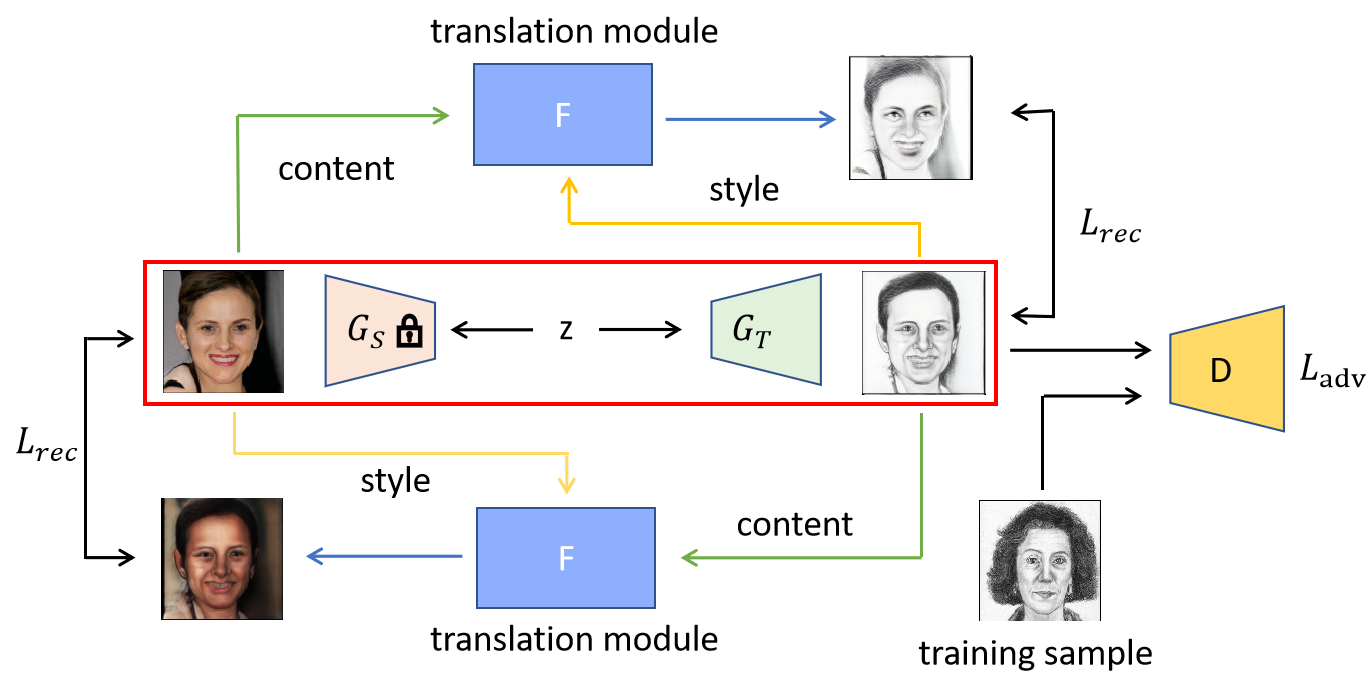} \\
  \caption
    { Overview of the proposed framework. Our approach uses a translation module $F$ for content preservation and a discriminator $D$ for style adaptation. $F$ will take a pair of images (shown in the red box) generated with the same latent code as input, and try to reconstruct them. We then use the reconstruction loss $L_{rec}$ to encourage these paired images to have the same content.}
    \label{lp}
  \label{fig:frame}
\end{figure}

\section{Related work}
\noindent
 \textbf{Few-shot image generation.} Few-shot image generation aims to generate diverse and realistic images with limited training data. A popular way to do this is to adapt a source model pre-trained on sufficient data of source domain to the target domain with few training data. Due to the great fitting ability of GAN model, the training can easily overfit to the training samples. To address this, many fine-tune based methods \cite{noguchi2019image,robb2020few,zhao2020leveraging,wang2020minegan} have been proposed. However, most of them still needs a relative large amount of data (more than 100) and fail to produce high-quality images. Recently, Ojha et al. \cite{ojha2021few} and Xiao et al. \cite{xiao2022few} propose to maintain a prior correspondence such as cross domain correspondence and correlation consistency loss to supervise training. Different from these works, our method applies an image translation module trained with the GAN model to supervise training without any prior knowledge.

 \noindent
 \textbf{Image to image translation.}  An alternative perspective of our training process can be seen as an image-to-image translation, where we aim to convert an image generated by the source model to the target domain. Then it is natural to only change the style of the image while preserve its content. An intuitive way to do this is to directly apply arbitrary style transfer methods such as AdaIN \cite{huang2017arbitrary}. However, the ”style” in these methods is not defined by the source and target domains, they refer more to the low-level semantic information extracted by some pre-trained classification network \cite{gatys2016image}. Therefore, as is shown in Fig \ref{fig:sketch}, these methods are likely to fail to properly transfer the domain-relevant style.

Image to image translation focuses on converting images to another domain. However, these methods \cite{liu2017unsupervised,zhu2017unpaired} are not designed for few-shot settings and require large amount of data from both source and target domain. Liu et al. \cite{liu2019few,saito2020coco} has proposed a framework to extract style and content information to address limited data from target domain. However, the content and style extractor still needs to be trained on sufficient labeled data (data with different class labels), which is not available in our case as the source domain is unlabeled.
% To overcome this limitation, we devised a strategy to generate training data using our generative models: given the same latent code, the target and source generative models can produce a pair of images from different domains, which makes a translator possible to train.  
% Notably, in their work \cite{liu2019few}, Liu et al. applied several methods to address the unpaired image translation since there is no two animals of different domains are at exactly the same pose, which happens not to be the case in our setting: we do have two images from different domains with the same content: the two images generated by $G_S$ and $G_T$ with a same latent code $z$. This makes it great easier for our translation module's training.

\section{Approach}
We are given a source generator $G_S$, which is trained on a large unlabeled dataset of source domain. We want to use $G_S$ and a few training samples (10 samples) to train a target generator $G_T$. The goal of our training is to transfer the rich content context of $G_S$ to $G_T$ while adapting the style context generated by $G_T$ to the target domain. To begin with, we initialize the weight of $G_T$ to $G_S$.

\begin{figure*}[t]
\centering
    \includegraphics[width=18.3cm]{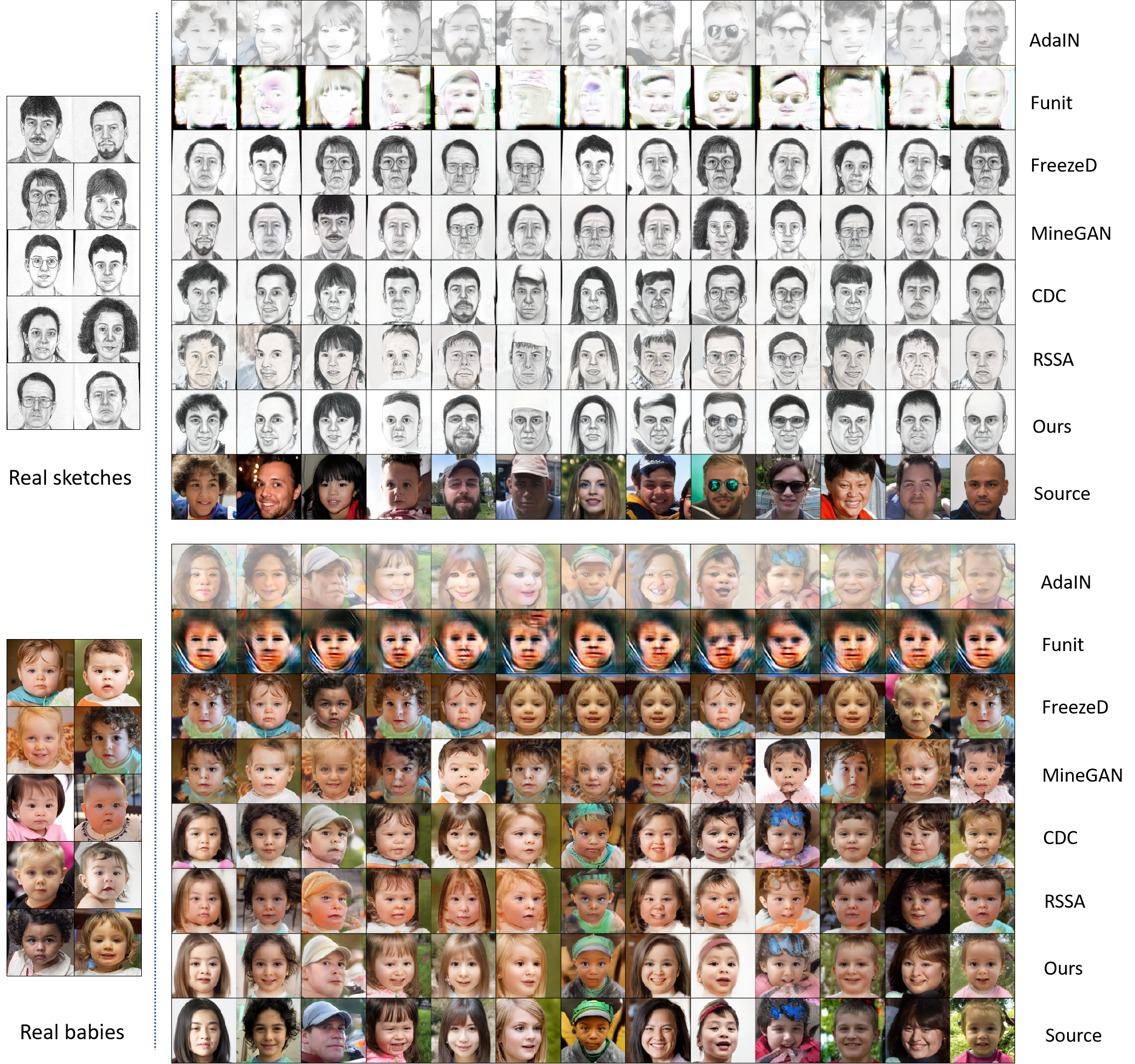} \\
  \caption
    { Comparison results with different baselines on \textit{FFHQ} $\rightarrow$ \textit{Sketches} and \textit{FFHQ} $\rightarrow$ \textit{FFHQ-babies}. For AdaIN , we randomly choose a real image as the style image. For other GAN-based methods, we keep the latent code same (across columns). Our method generates results of higher quality and diversity which better correspond to the source domain images.}
    \label{lp}
  \label{fig:sketch}
\end{figure*}

We assume that, in an ideal training process, \textbf{with same latent code $z$, images generated by $G_T$ and $G_S$ should share the same content}. This is because it should be easier for the generator $G_T$ to only change the style of an image than doing extra modification. However, due to the limited training data, the target generator is hard to distinguish between content and style, which therefore leads to overfitting (Fig \ref{fig:sketch}). 

As a result, we introduce a translation module to help the model learn the knowledge about style and content. To achieve this, we apply a paired image reconstruction procedure (Fig \ref{fig:frame}) to encourage the content preservation, every epoch of training contains three steps, where $D$, $G_T$ and $F$ are trained separately. 
% \begin{itemize}[leftmargin=.5in]
%   \item [Step 1] 
%   Use adversarial loss $L_{adv}$ to train $G_T$ and $D$, meantime, use reconstruction loss $L_{rec}$ to regularize $G_T$'s training.
%   \item [Step 2]
%   Use reconstruction loss ${L^{'}}_{rec}$ to train $F$ with paired images generated by $G_S$, $G_T$ and images translated by the translator.
% \end{itemize}

The architecture of our translation module is interpreted in Sec. \ref{trans}; the trade-off between style-adaptation and content preservation is discussed in Sec. \ref{Reg}, where we also explain our  content preservation approach. Finally, we explain our learning objective and applied losses in Sec. \ref{learn}.
\begin{figure}[t]
\includegraphics[width = \linewidth,scale = 1.0]{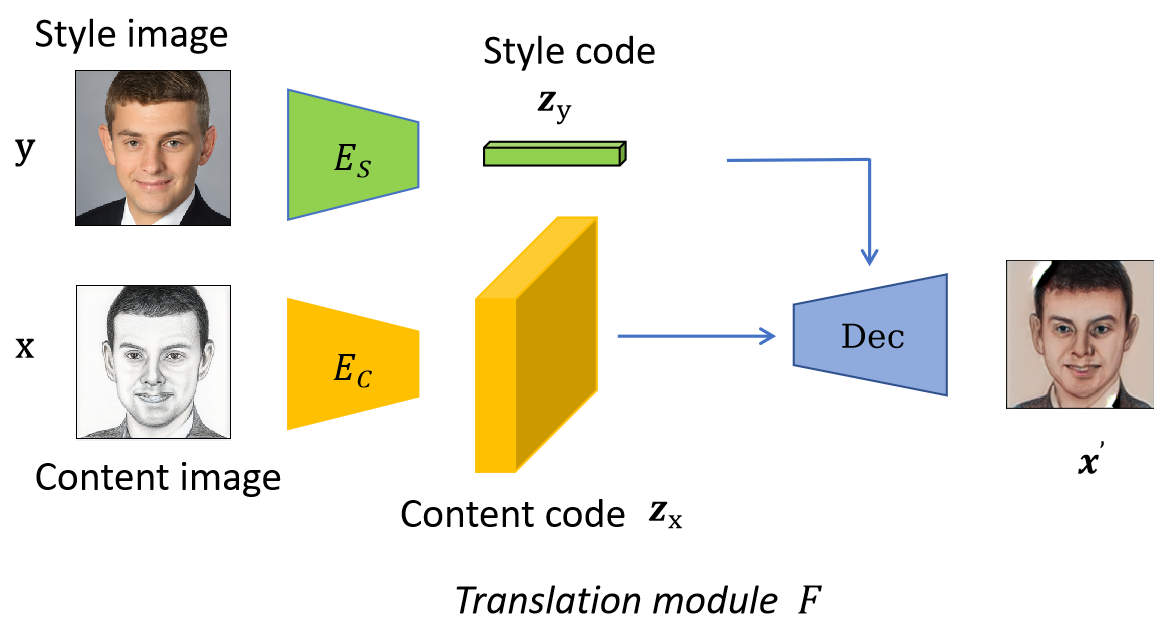} \\
  \caption
    { Architecture of the proposed translation module $F$. $F$ consists of a style encoder $E_S$, a content encoder $E_C$ and a decoder $Dec$. To generate a translation output $x^{'}$, $F$ combines the style code $z_y$ extracted from the input style image $y$ with the content code $z_x$ extracted from the input content image $x$. }
  \label{fig:trans}
\end{figure}

\begin{figure}[t]
\includegraphics[width = \linewidth,scale = 1.0]{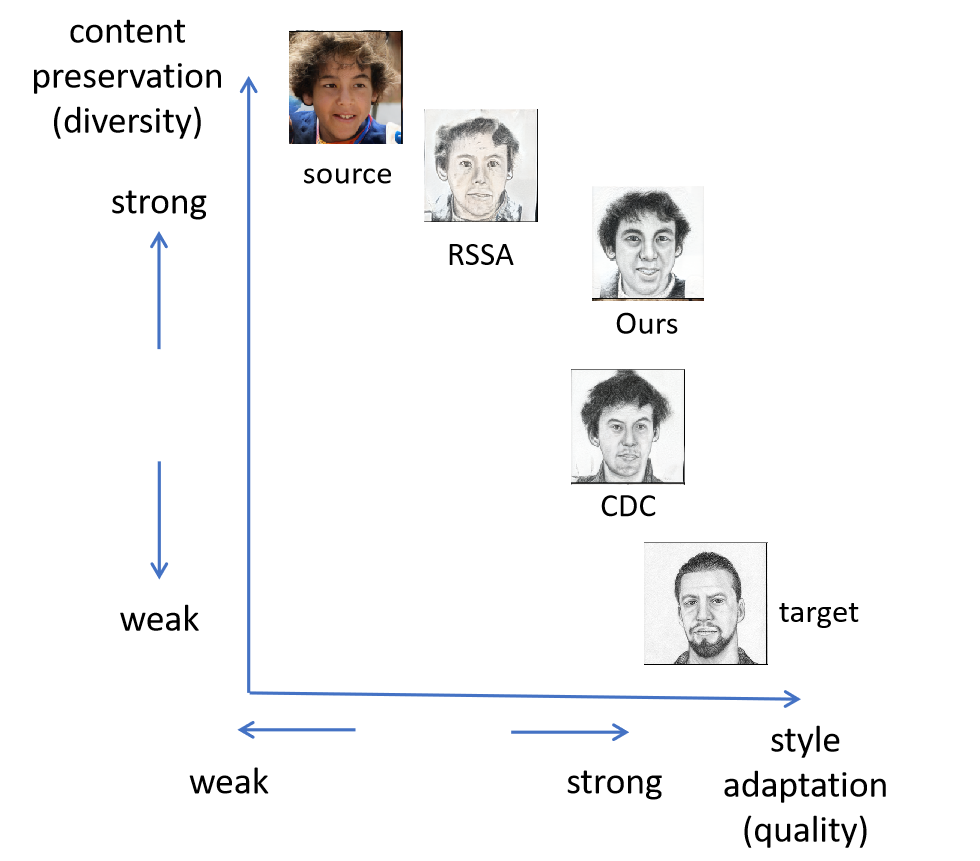} \\
  \caption
    { Schematic diagram of the trade-off. The horizontal axis represents style adaptation level (represents image quality), and the vertical axis represents content preservation (represents image diversity). The upper left corner is the output of the source model, and the lower right corner is the training example of the target domain. }
  \label{fig:trade}
\end{figure}

\begin{figure*}[t]
    \includegraphics[width=18.3cm]{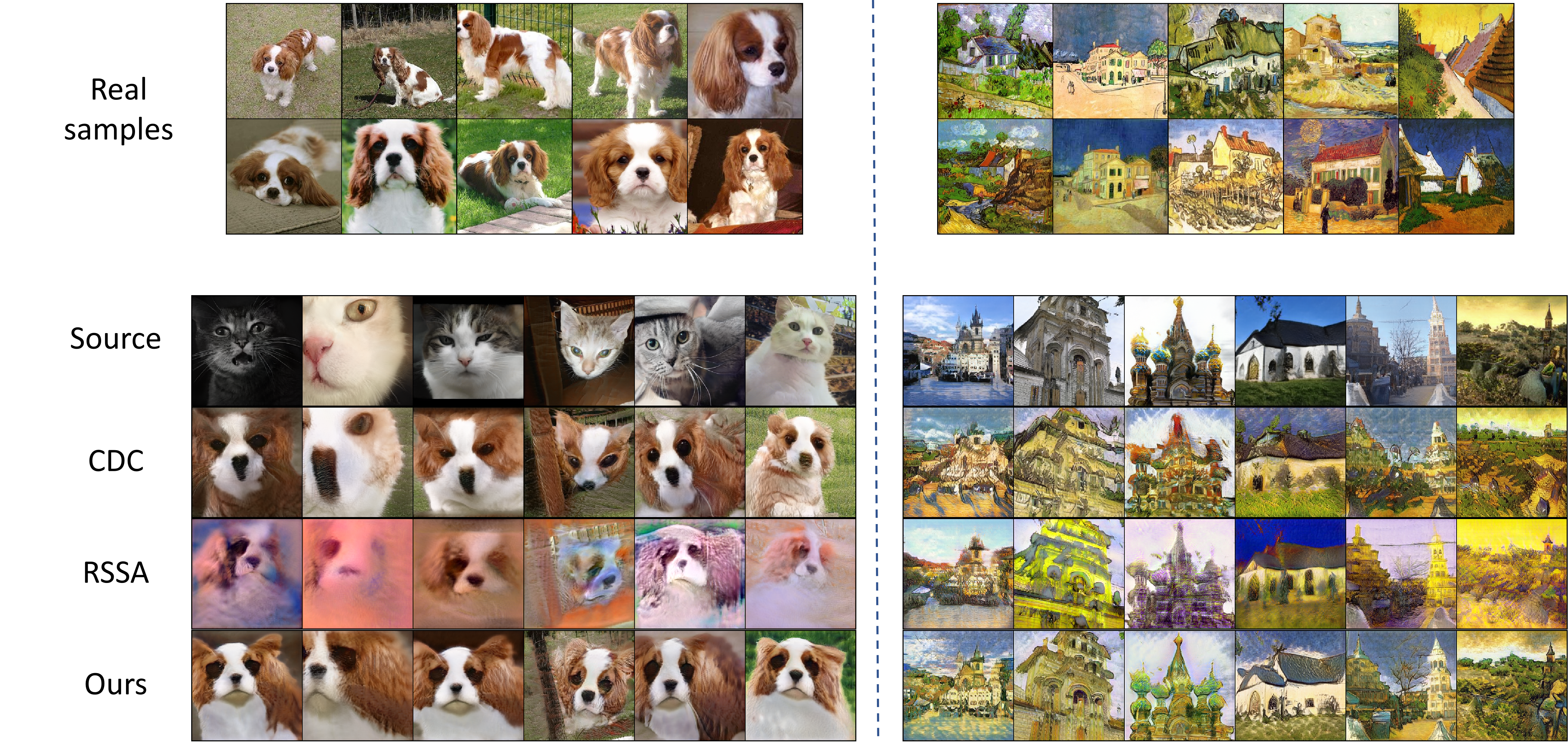} \\
  \caption
    { Comparison results with CDC and RSSA on \textit{LSUN Cats} $\rightarrow$ \textit{LSUN Spaniels} and \textit{LSUN Churches} $\rightarrow$ \textit{Van Gogh Houses}.}
    \label{lp}
  \label{fig:spaniel_vangogh}
\end{figure*}

\subsection{Image translation module}
\label{trans}
As is shown in Fig \ref{fig:trans}, our image translation module consists of a content encoder $E_C$, a style encoder $E_S$ and a decoder $Dec$. The content encoder $E_C$ maps the input content image $x$ to a content code $z_x$, the style encoder $E_S$ maps the input style image $y$ to a style code $z_y$. The decoder $Dec$ takes these two codes as input, and generates an image that combines the corresponding content and style. $Dec$ consists of a couple of adaptive instance normalization layers, which will use $z_y$'s mean and variance to normalize $z_x$, the convolutional layers will then upscale it to the final output image.

\subsection{The trade-off between style adaptation and content preservation}
\label{Reg}
From another point of view, the few-shot generative model adaption can be seen as a trade-off between style adaptation and content preservation. The style adaptation is naturally done by a discriminator, we therefore need to find a proper way to preserve content (diversity). The fine-tune based methods such as FreezeD \cite{mo2020freeze} focus little on this and thus end up overfitting. 

On the other hand, CDC \cite{ojha2021few} and RSSA \cite{xiao2022few} try to preserve content by maintaining a correspondence between two images, as is shown in Fig \ref{fig:method_compare}. The authors used these correspondences to estimate the true \textit{content}. The drawback of this is that the quality of this estimation varies with different kinds of \textit{source} $\rightarrow$ \textit{target} adaptations. As depicted in Fig \ref{fig:trade}, \ref{fig:sketch}, we find that CDC is weak in content preservation, as it captures relatively little content context from the source domain. In contrast, RSSA overemphasizes content preservation and is prone to preserving lots of related information that may not be desired for certain target domains (e.g. when we transfer from FFHQ to sketches, some of the output will have color). Moreover, the trade-off between content preservation and style adaptation is very dynamic, as demonstrated in Fig \ref{fig:sketch}, when we transfer FFHQ to babies, neither CDC nor RSSA can preserve diverse mouth poses. We will further discuss this with qualitative and quantitative results in Sec. \ref{sec:performance}.

Despite their drawbacks, it is worth noting that in several \textit{source} $\rightarrow$ \textit{target} adaptations, CDC and RSSA have achieved relatively good results. This suggests that there are different proper trade-off points for different adaptations, which can be achieved by estimating the content information in a suitable way. When the estimated content information is appropriate for the given \textit{source} $\rightarrow$ \textit{target} adaptation, CDC or RSSA can achieve good results by balancing between content preservation and style adaptation.

Accordingly, we need a dynamic strategy to find suitable trade-offs for different target domains. Instead of using pre-defined correspondence losses, we introduce a translation module to learn to separate content and style.  Specifically, we propose to apply paired image reconstruction to preserve the content information during the adaptation process, as is shown in Fig \ref{fig:frame}. Given a latent code $z$, $G_T$ and $G_S$ will generate a pair of images $x$ and $y$ from different domains. We use the translation module $F$ to reconstruct $y$ from the content of $x$ and the style of $y$. When $F$ is frozen, the reconstruction loss encourages $x$ to have the same content as $y$ for if there is any modification to the content of $x$, the translation module would fail to find the original content of $y$. Similarly, we can let $F$ take the content of $y$ and the style of $x$ to reconstruct $x$ according to the symmetry.

During experiment, we find that the $l_1$ loss equally takes every pixel into account, which makes reconstruction harder. On the other hand, LPIPS loss \cite{zhang2018unreasonable} measures the deep features of two images, which makes the training much faster and more stable. $L_{rec}$ is then given by:
 % \begin{equation}
 \begin{align}
     L_{rec} =\; & \mathbb{E}_{z \sim p(z)}[LPIPS(F(G_T(z),G_S(z)),G_S(z)) \\&+ LPIPS(F(G_S(z),G_T(z)),G_T(z))]
 \end{align}
 % \end{equation}

\subsection{Learning}
\label{learn}
We train the whole framework by solving the given optimization objective :
\begin{equation}
    \min_{G_T,F} \max_D L_{adv}(D,G_T) + \lambda_1L_{rec}(G_T) + \lambda_2{L^{'}}_{rec}(F)
\end{equation}
where $L_{adv}$ refers to the GAN loss, $L_{rec}$ and ${L^{'}}_{rec}$ refer to the reconstruction loss used for target generator $G_T$ and translator $F$ respectively.
The GAN loss is an adversarial loss given by:
\begin{equation} 
\begin{aligned}
    L_G &= - \mathbb{E}_{z \sim p(z)}[log(D(G_T(z))] \\
    L_D &= \mathbb{E}_{x \sim D_t}[log(1 - D(x)] + \mathbb{E}_{z \sim p(z)}[log(D(G_T(z))]
\end{aligned}
\end{equation}

For $L_D$, we follow the idea of \cite{ojha2021few} to use a combination of image-wise and patch-wise discriminator loss.

The reconstruction loss ${L^{'}}_{rec}$ is designed to train the translator $F$. Similar to previous step explained in Sec. \ref{Reg}, $G_T$ and $G_S$ can generate a pair of images with same content and different style. We freeze $G_T$ to let $F$ learn to reconstruct one image with the other's content and its own style. 

 In order to ensure that the style and content of an image could cover all its information, we also let $F$ perform self reconstruction, where the input content image and style image would be the same.
 ${L^{'}}_{rec}$ is then given by:
 % \begin{equation}
 \begin{align}
     {L^{'}}_{rec} =\; & \mathbb{E}_{z \sim p(z)}[LPIPS(F(G_T(z),G_S(z)),G_S(z))\\
     &+ LPIPS(F(G_S(z),G_T(z)),G_T(z))\\
     &+ LPIPS(F(G_S(z),G_S(z)),G_S(z))\\
     &+LPIPS(F(G_T(z),G_T(z)),G_T(z))]
 \end{align}
 % \end{equation}

We find that when the translation module is sufficiently trained in every iteration, the training process would be fairly stable. For most adaptations, we train the generator and the discriminator once, and the translation module 4 times in each iteration of training. Additional training details can be found in the supplementary.

\begin{figure*}[t]
    \includegraphics[width=18.3cm]{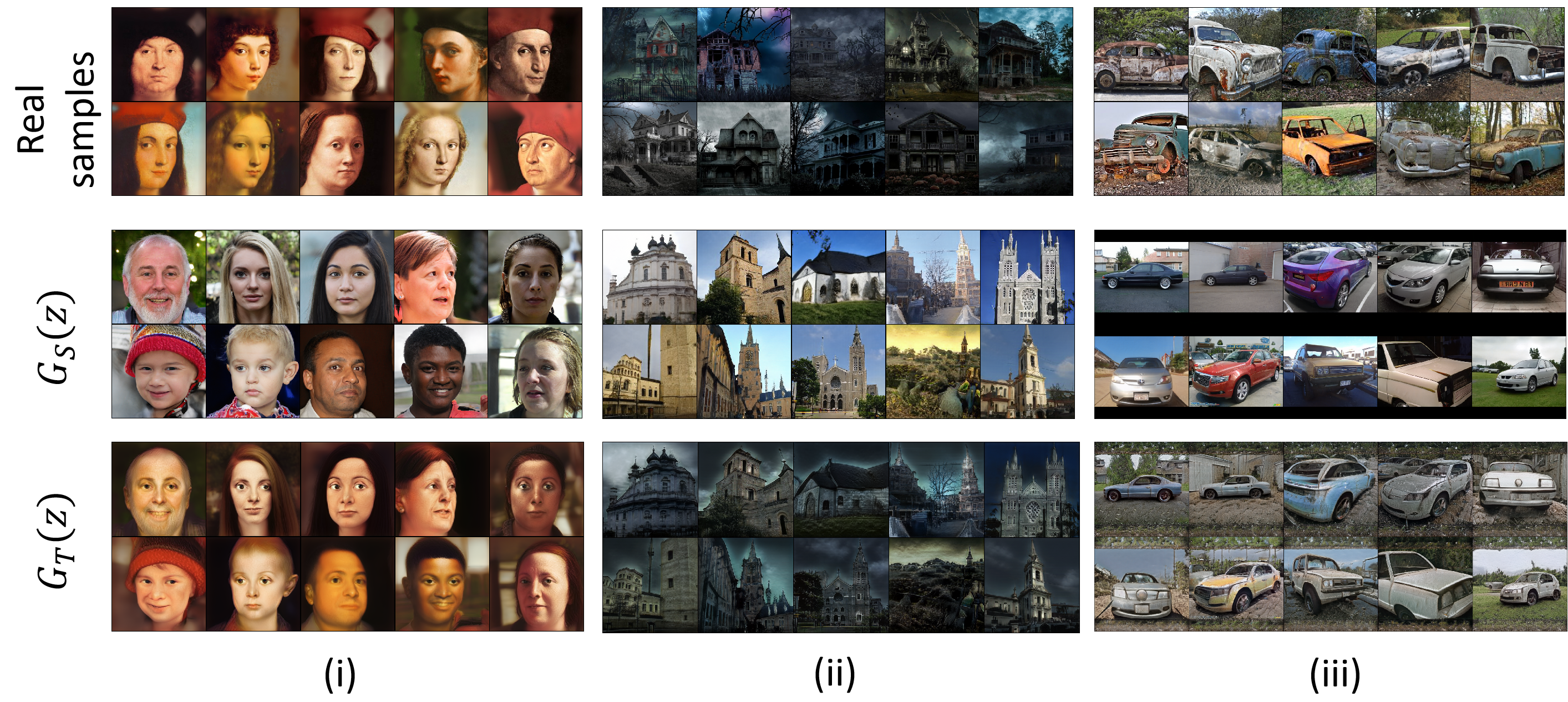} \\
  \caption
    { Results of different adaptation settings: $(i)$ \textit{FFHQ} $\rightarrow$ \textit{Raphael Paintings} ; $(ii)$ \textit{LSUN Churches} $\rightarrow$ \textit{Haunted Houses}; $(iii)$ \textit{LSUN Cars} $\rightarrow$ \textit{Wrecked Cars}.}
    \label{lp}
  \label{fig:other}
\end{figure*}

\begin{figure*}[t]
    \includegraphics[width= 18cm]{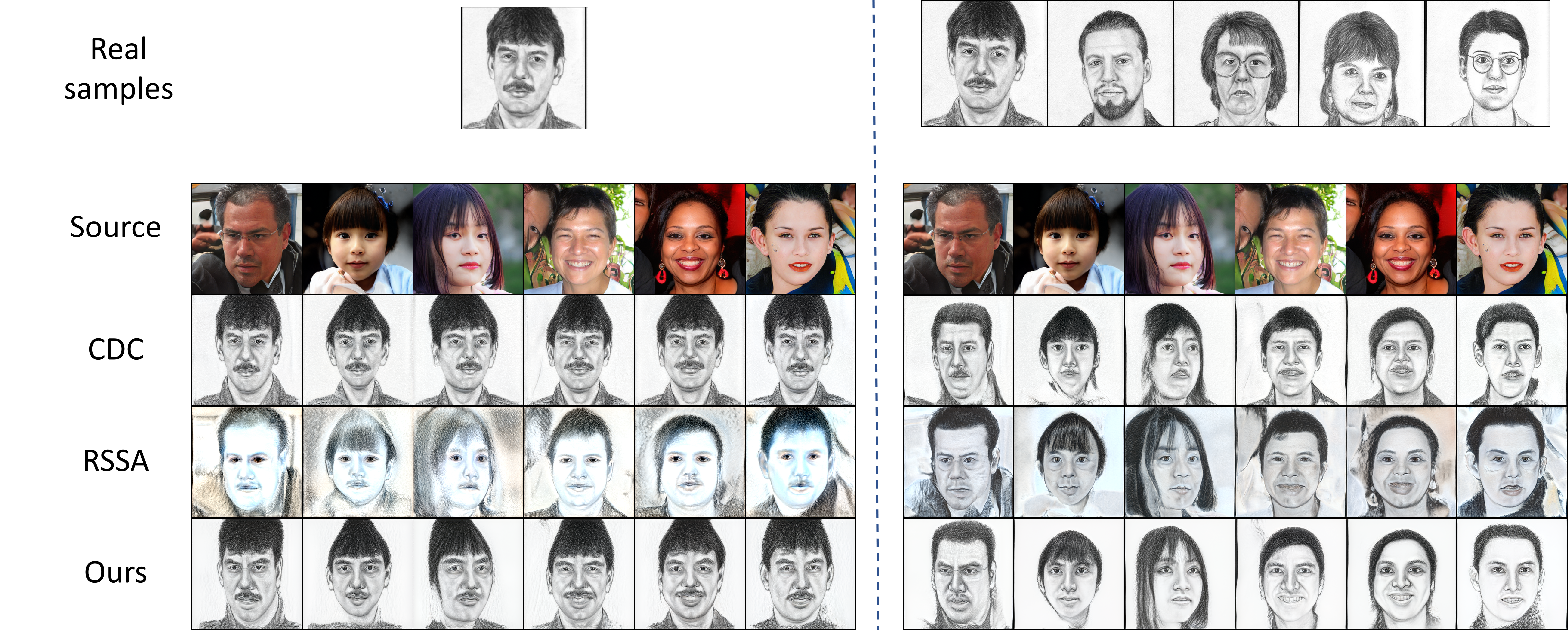} \\
  \caption
    { Comparison results with CDC and RSSA on \textit{FFHQ} $\rightarrow$ \textit{Sketches} in 1-shot and 5-shot settings.}
  \label{fig:sketch5}
\end{figure*}

\begin{figure*}[t]
    \includegraphics[width= 18cm]{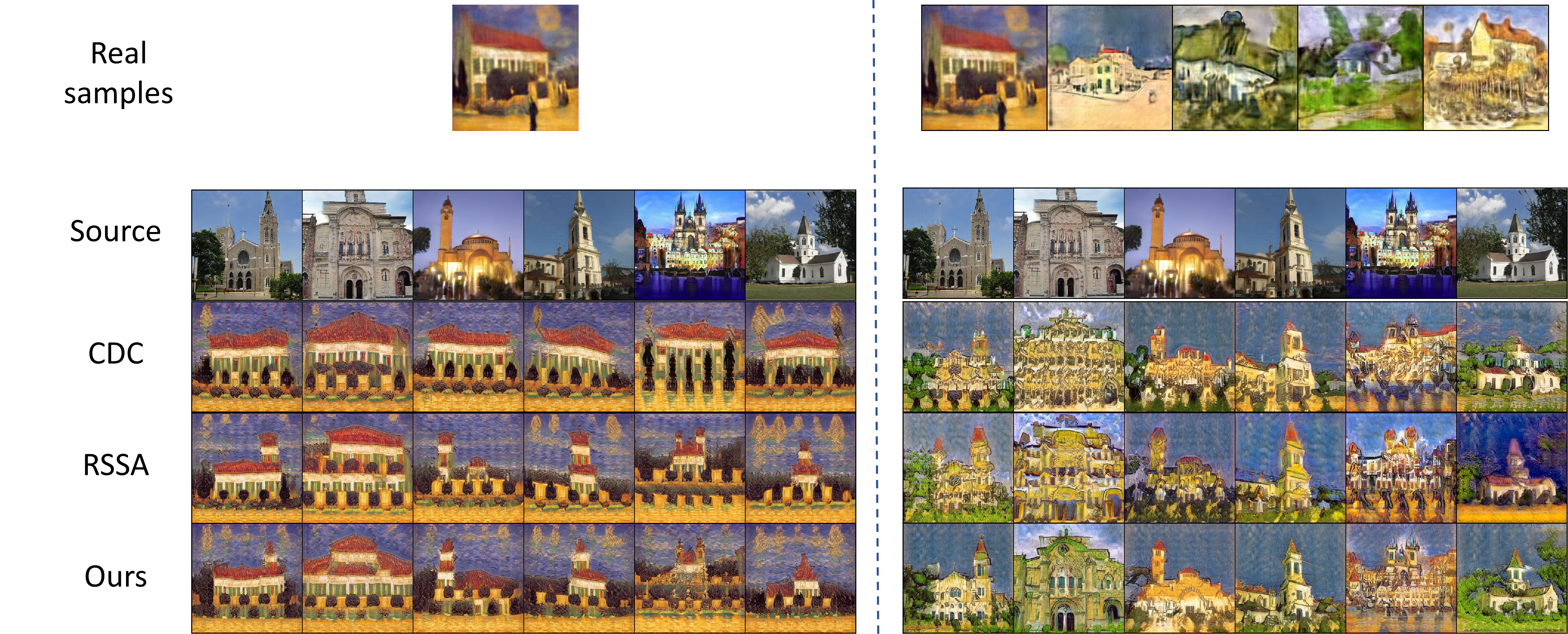} \\
  \caption
    { Comparison results with CDC and RSSA on \textit{LSUN Churches} $\rightarrow$ \textit{Van Gogh Houses} in 1-shot and 5-shot settings.}
  \label{fig:vangogh5}
\end{figure*}

\begin{figure*}[t]
    \includegraphics[width= 17cm]{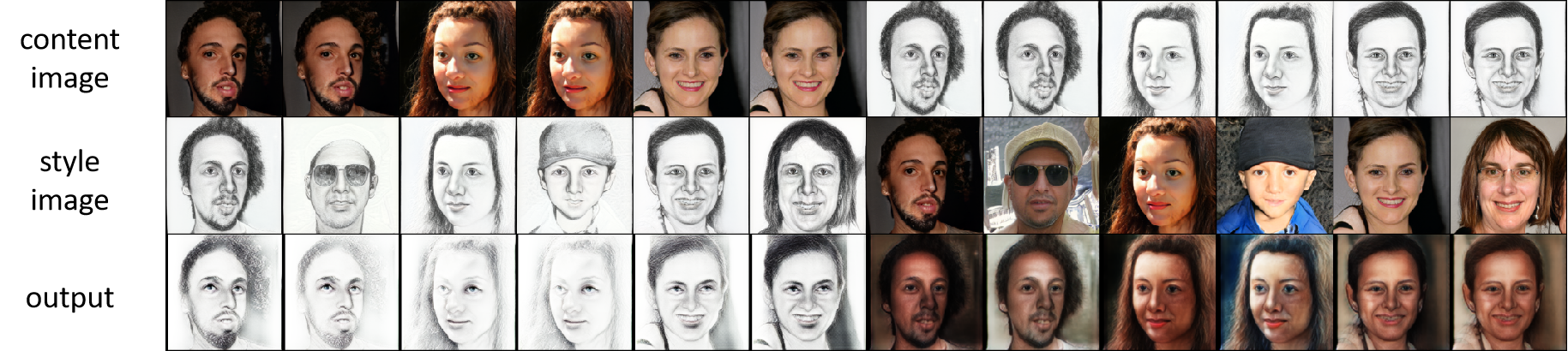} \\
  \caption
    { The results of the translation module after training. The module takes the first row as content image, and the second row as style image. The outputs show that after training the translation module is capable to separate style and content and translate images from source domain to target domain.}
    \label{lp}
  \label{fig:transl}
\end{figure*}
\begin{figure}[t]
    \includegraphics[width=\linewidth]{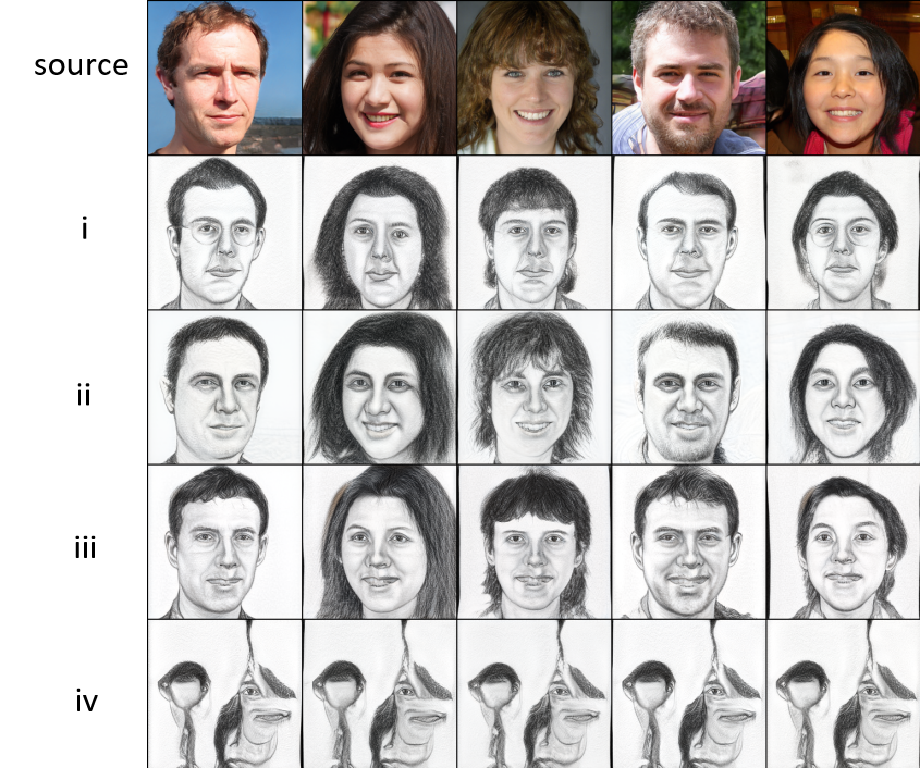} \\
  \caption
    { Comparison results with different losses on \textit{FFHQ} $\rightarrow$ \textit{Sketches}. $(i)$ compute the $l1$ loss of the original images and reconstructed images; $(ii)$ compute the $LPIPS$ loss of the original images and reconstructed images; $(iii)$ compute the $l1$ loss on the content codes of the content images and reconstructed images, and the $l1$ loss on the style codes of the style images and reconstructed images; $(iv)$ put the reconstructed image into the corresponding discriminator and use the adversarial loss as reconstruction loss.}
    \label{lp}
  \label{fig:loss}
\end{figure}

\begin{figure}[t]
    \includegraphics[width=\linewidth]{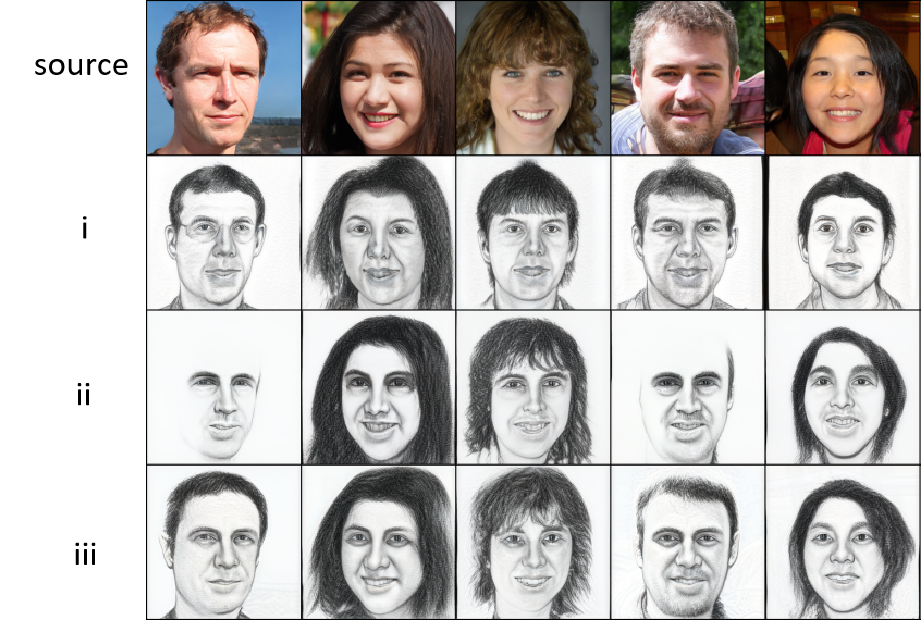} \\
  \caption
    { Comparison results with different reconstruction choices on \textit{FFHQ} $\rightarrow$ \textit{Sketches}. $(i)$ reconstruct the source image only; $(ii)$ reconstruct the target image only; $(iii)$ reconstruct both the source image and the target image. }
    \label{lp}
  \label{fig:recon}
\end{figure}
\section{Experiment}
In this section, we discuss the effectiveness of 
our approach in several few-shot settings. We compare our method qualitatively and quantitatively with the following few-shot image generation baselines: FreezeD \cite{mo2020freeze}, MineGAN \cite{wang2020minegan}, CDC \cite{ojha2021few} and RSSA \cite{xiao2022few}. 
We use the StyleGANv2 \cite{karras2020analyzing} models pre-trained on sour different datasets: $(i)$ Flickr-Faces-HQ (FFHQ) \cite{karras2019style}, $(ii)$ LSUN Churches \cite{yu2015lsun}, $(iii)$ LSUN Cars \cite{yu2015lsun} and $(iv)$ LSUN Cats \cite{yu2015lsun}.

We adapt the source GAN models to various target domains including: $(i)$ face sketches \cite{wang2008face}, $(ii)$ FFHQ-babies \cite{karras2019style}, $(iii)$ haunted houses \cite{ojha2021few}, $(iv)$ LSUN spaniels \cite{yu2015lsun}, $(v)$
village painting by Van Gogh \cite{ojha2021few},$ (vi)$ wrecked/abandoned cars \cite{ojha2021few} and $(vii)$ Raphael paintings \cite{ojha2021few}.

\subsection{Performance evaluation}
\label{sec:performance}
\noindent
\textbf{Qualitative comparison.} Fig \ref{fig:sketch} shows the results of different methods on two transfer settings. We can observe that directly style transfer (AdaIN \cite{huang2017arbitrary}) fails the transfer task for it only replaces the low-level semantic information of the output images and fails to capture the real domain-relevant features. Direct image-to-image translation (FUNIT \cite{liu2019few}) is also prone to collapse in these settings. Without a labelled source domain that contains enough source classes, the translator would find it very hard to separate style and content information between source domain and a very small target domain. As for fine-tune based few-shot GAN adaptation methods (FreezeD \cite{mo2020freeze}, MineGAN \cite{wang2020minegan}, the output images strongly overfit to the reference images of the target domain. Though these methods perform well with hundreds of training samples, they are ineffective in handling extremely few-shot settings. 

On the other hand, CDC \cite{ojha2021few} and RSSA \cite{xiao2022few} manage to generate diverse and realistic images compared to other baselines. However, there are still some issues with their correspondence losses. For CDC, we can see that the generated sketch images are not quite identical to the source images. This indicates that CDC may lose some diversity of facial expressions when transferred to the target domain. On the contrary, RSSA can preserve visual attributes well in sketches, but sometimes overly restricts the output. This over-restriction can even result in some output images having colors, which should not be the case for sketches. Besides, RSSA does not handle light and shadow very well, resulting in unnatural spots on the face. Moreover, despite the strong effect RSSA has archieved in \textit{FFHQ} $\rightarrow$ \textit{Sketches} adaptation, it can not provide enough constraints in \textit{FFHQ} $\rightarrow$ \textit{FFHQ-babies} adaptation, where neither CDC nor RSSA can preserve diverse mouth poses. They can even have some strange distortions in such settings. This indicates that the assumed correspondence losses cannot properly balance style adaptation and content preservation.

Our method achieves best results in these two adaptations. As depicted in Fig \ref{fig:sketch}, in \textit{FFHQ} $\rightarrow$ \textit{Sketches} adaptation, our output images can better capture the diverse facial expressions without bringing color and can better address the light and shadow issues; in \textit{FFHQ} $\rightarrow$ \textit{FFHQ-babies} adaptation, our methods can preserve refined details such as the mouth poses and does not output unnatural distortions. 
We further compare our method with CDC and RSSA on \textit{LSUN Cats} $\rightarrow$ \textit{LSUN Spaniels} and \textit{LSUN Churches} $\rightarrow$ \textit{Van Gogh Houses} adaptations, as is shown in Fig \ref{fig:spaniel_vangogh}. We can see that RSSA fails to address the adaptation between two relatively distant domains such as \textit{LSUN Cats} $\rightarrow$ \textit{LSUN Spaniels}. Even between a more related pair of domains such as \textit{LSUN Churches} $\rightarrow$ \textit{Van Gogh Houses}, there still exists undesired distortion in RSSA's outputs. On the other hand, CDC can generate acceptable results in these two settings. Our method outperforms both of them, indicates that our method can properly balance the style adaptation and content preservation on various \textit{source} $\rightarrow$ \textit{domain} adaptations.

In Figure \ref{fig:other}, we show more results with different source $\rightarrow$ target adaptation settings. Supervised by paired image reconstruction loss, the generator can produce images that successfully preserve diversity from the source domain while the style fits the target domain.
\begin{table}[t]
    \caption
    {FID scores ($\downarrow$) for domains with abundant data. Our method achieves the lowest FID score, demonstrating its superiority in adapting the style to the target domain. In contrast, RSSA shows a very high FID score, indicating that it is weak in terms of style adaptation.}
    \centering
  \begin{tabular}{l|c|c|c} 
  \toprule
  \linespread{2}
     & babies & sketches & spaniels \\ \hline \midrule
    FreezeD  &113.01  &50.53 & 88.53 \\
    MineGAN & 100.15  & 69.77 &79.12 \\
    CDC & 77.07  & 47.33  & 65.74\\
    RSSA  &138.25   & 84.73 & 160.89 \\
    Ours& \textbf{70.50} &\textbf{45.01}& \textbf{62.74} \\\bottomrule
  \end{tabular}
  \label{table:fid}
\end{table}

\begin{table}[t]
\caption
    { Intra-cluster pairwise LPIPS distance ($\uparrow$). CDC shows relatively lower distance (less diversity), suggesting its weakness in content preservation. }
\centering
  \begin{tabular}{l|c|c|c} 
  \toprule
  \linespread{2}
     & babies & sketches & spaniels \\ \hline \midrule
    CDC & 0.567 $\pm$ 0.019  & 0.400 $\pm$ 0.020  & 0.663 $\pm$ 0.023\\
    RSSA  &0.575 $\pm$ 0.020   & 0.514 $\pm$ 0.012 & 0.670 $\pm$ 0.034 \\
    Ours& 0.576 $\pm$ 0.013 &0.436 $\pm$ 0.033& 0.667 $\pm$ 0.040 \\\bottomrule
  \end{tabular}
  \label{table:lpips}
\end{table}

\begin{table}[t]
 \caption
    {Balance index ($\uparrow$) for quality and diversity.}
\centering
  \begin{tabular}{l|c|c|c} \toprule
  \linespread{2}
     & babies & sketches & spaniels \\ \hline \midrule
    CDC & 7.36  & 8.45  & 10.09\\
    RSSA  &4.16   & 6.07 & 4.16 \\
    Ours& \textbf{8.17} &\textbf{9.67}& \textbf{10.63} \\\bottomrule
  \end{tabular}
  \label{table:balance}
\end{table}
\noindent
\textbf{Quantitative comparison.} 
% We use two Most of our experiments are done on a very small target domain, which cannot provide enough data for quality evaluation. 
We use three datasets with abundant data that meet the requirement of evaluation: the original Sketches, FFHQ-babies, and LSUN-spaniels datasets, which roughly contain 300, 2500 and 200 images respectively. We let the models trained with different methods to generate $5000$ samples, which are used to calculate the FID score for each method.

Table \ref{table:fid} shows the FID \cite{heusel2017gans} score of different methods. Our method achieves the best fid score on the three datasets, indicating that our method generates images that best model the true distribution of target domain.  However, the FID score
would not reflect the degree of overfitting.

Therefore, we use the intra-cluster pairwise LPIPS distance \cite{ojha2021few} to measure the diversity level. As is shown in Table \ref{table:lpips}, our method consistently achieves higher average LPIPS distance than CDC. On the other hand, although RSSA achieves higher LPIPS distance in \textit{FFHQ} $\rightarrow$ \textit{Sketches} and \textit{Cats} $\rightarrow$ \textit{Spaniels} adaptations, it gets a very high FID scores in these adaptations (meaning the learned distribution is very different from the target domain). This is consistent with our previous analysis: CDC is relatively weak in content preservation, which fails to generate more diversity (has the lowest LPIPS distance) while RSSA overemphasizes the diversity and fails to generate similar distribution to the target domain (has very high FID score).

To assess the combined performance of the model in terms of diversity and quality more clearly, we propose a balance metric by incorporating FID score(FID) and LPIPS distance(LD) as follow:
 \begin{equation}
     balance =\frac{100\cdot LD}{FID}
 \end{equation}
 Higher score means better performance in balancing quality and diversity. The comparison results are shown in Table \ref{table:balance}, we can see that our method outperforms CDC and RSSA.

\subsection{Ablation study}
\noindent
\textbf{Effect of target dataset size.} We further explore the effectiveness of our method compared to CDC and RSSA in 1-shot and 5-shot settings.

Fig \ref{fig:sketch5} shows the results on \textit{FFHQ} $\rightarrow$ \textit{Sketches} adaptation. For the 1-shot setting, we can observe that RSSA is so strong at content preservation that with only one examples, the outputs can still resemble the corresponding images from the source domain. However, it falls short in style adaptation as it introduces color that is not consistent with the sketch style. On the other hand, CDC and our method just output similar faces with different poses. Our method can generate more diverse poses than CDC, such as the poses of mouth. The evaluation of 1-shot generation is very vague, for it is very hard to tell what is style and content.  The results of 5-shot setting are similar to those of 10-shot setting. We can see that the results of CDC are not very identical to the corresponding images from the source domain, indicating its shortness in content preservation. On the other hand, RSSA can generate outputs resembling the original images but brings some undesired color, suggesting its weakness in style adaptation. Our method can generate diverse outputs without bringing any color.

Fig \ref{fig:vangogh5} shows the results on \textit{LSUN} $\rightarrow$ \textit{Van Gogh Houses} adaptation. For the 1-shot setting, we can observe that our method and RSSA can generate churches with different shapes while CDC can only output similar churches. For 5-shot setting, we can see that both CDC and RSSA have unnatural distortions while our method can generate churches similar to the original images and successfully adapts the style to Van Gogh paintings.

\noindent
\textbf{The translation module.} We examine the output of the translation module to evaluate its function. As is shown in Fig \ref{fig:transl}, after
training the translation module is capable to separate style and content and translate images from source domain to target domain.

\noindent
\textbf{Choice of reconstruction loss.} We explore four kinds of reconstruction loss in our experiments: $(i)$ compute the $l1$ loss of the original images and reconstructed images; $(ii)$ compute the $Lpips$ loss of the original images and reconstructed images; $(iii)$ compute the $l1$ loss on the content codes and the style codes; $(iv)$ put the reconstructed images into the corresponding discriminator and use the adversarial loss as reconstruction loss. The result is shown in Fig \ref{fig:loss}. We can see that putting $l1$ loss on two images results in overfitting, probably because it overemphasizes the pixel similarity and makes it great harder for the translator to do the reconstruction; on the other hand, $LPIPS$ loss can relax the translator, let it focus on reconstruct meaningful information and thus has the best results. Putting $l1$ loss directly on the style and content code can generate similar results as $(ii)$, but will loss some diversity. Using discriminator loss for reconstruction fails to produce valid results.

\noindent
\textbf{Choice of reconstruction method.} As is mentioned in Sec. \ref{Reg}, we actually have three choices for reconstruction: $(i)$ reconstruct the source images only; $(ii)$ reconstruct the target images only; $(iii)$ reconstruct both the source images and the target images. The results are shown in Fig \ref{fig:recon}. We can observe that if we only reconstruct target images, some edges of the face will appear blank; while only 
 reconstructing source images doesn't have this issue. Both of them have overfitting issues and cannot fully capture the face expression details. On the other hand, by reconstructing both source and target images, the face will become more refined and more identical to the source images.

%-------------------------------------------------------------------------
\section{Conclusion and limitation}
In this paper, we propose a novel content preservation method to address the image generation problem in extremely few shot setting. With the help of the paired image reconstruction, the generative model can learn to preserve rich content context inherited from the source domain. Due to the flexible trade-off strategy between style adaptation and content preservation, our approach can successfully generate diverse and realistic images under various \textit{source} $\rightarrow$ \textit{domain} adaptation settings.

\noindent
\textbf{Limitations.} Despite the compelling results our method achieves, there are still some limitations. Firstly, there is still some overfitting on certain details of the output. For example, in \textit{FFHQ} $\rightarrow$ \textit{Sketches} adaptation, the generated sketch faces will tend to show a little teeth even if the mouth is initially closed on the source image. In \textit{FFHQ} $\rightarrow$ \textit{FFHQ-babies} adaptation, the hair color is a little shallower than the corresponding source image. Besides, the adaptation should be conducted between similar domains for good results.

\bibliographystyle{IEEEtran}
\bibliography{egbib}
\end{document}